\documentclass[journal]{IEEEtran}
\usepackage{flushend}
\usepackage{epsfig}
\usepackage{epstopdf}
\AppendGraphicsExtensions{.tiff}

\usepackage{fancyhdr}
\usepackage{graphicx}
\usepackage[cmex10]{amsmath}
\usepackage{amssymb}
\usepackage{multirow}
\usepackage{rotating}
\usepackage[table]{xcolor}
\usepackage{threeparttable}
\usepackage{url}

\usepackage{color,soul}
\sethlcolor{yellow}
\usepackage{algorithmic}
\usepackage{algorithm}
\usepackage{amsthm}
\theoremstyle{definition}
 
\usepackage{verbatim} % for multi-line comments "\begin{comment}"
\begin{document}
%
% paper title
% can use linebreaks \\ within to get better formatting as desired
%\title{{\em AttentionBoost}: A New Boosting Algorithm that Learns What to Attend in Fully Convolutional Networks
\title{{\em AttentionBoost}: Learning What to Attend by Boosting Fully Convolutional Networks}
%Attention Learning by Boosting Fully Convolutional Networks
%
%What to Attend by Boosting Fully Convolutional Networks
%Boosting a Fully Convolutional Network to Learn Where to Attend
%to Learn What to Attend for Medical Image Segmentation
%
%
% author names and IEEE memberships
% note positions of commas and nonbreaking spaces ( ~ ) LaTeX will not break
% a structure at a ~ so this keeps an author's name from being broken across
% two lines.
% use \thanks{} to gain access to the first footnote area
% a separate \thanks must be used for each paragraph as LaTeX2e's \thanks
% was not built to handle multiple paragraphs
%

\author{Gozde~Nur~Gunesli, Cenk Sokmensuer, and Cigdem~Gunduz-Demir*,~\IEEEmembership{Member,~IEEE} \IEEEcompsocitemizethanks{\IEEEcompsocthanksitem 
%Manuscript received March 16, 2009; revised July 15, 2009 and August 30,
%2009. First published October 20, 2009; current version published February 17,
%2010. First published October 20, 2009; current version published February 17,
%2010. 
This work was supported by the Scientific and Technological Research Council of Turkey under the project number T{\"U}B\.{I}TAK 116E075.
\em{Asterisk indicates corresponding author.}
}
\thanks{G. N. Gunesli is with the Department of Computer Engineering, Bilkent University, Ankara TR-06800, Turkey (e-mail:
nur.gunesli@bilkent.edu.tr).}
\thanks{C. Sokmensuer is with the Department of Pathology, Medical School,
Hacettepe University, Ankara TR-06100, Turkey (e-mail: csokmens@hacettepe.edu.tr).}
\thanks{*C. Gunduz-Demir is with the Department of Computer Engineering and Neuroscience Graduate Program, Bilkent University, Ankara TR-06800, Turkey (e-mail: gunduz@cs.bilkent.edu.tr).}

%\thanks{Copyright (c) 2014 IEEE. Personal use of this material is permitted. However, permission to use this material for in any other purposes must be obtained from the IEEE by sending an email to pubs-permissions@ieee.org.}
%\thanks{Color versions of one or more of the figures in this paper are available online at htttp://ieeexplore.ieee.org}
%\thanks{Digital Object Identifier }
%Also, CGD acknowledges additional support from the Turkish Academy of Sciences Distinguished Young Scientist Award (TUBA GEBIP).

\thanks{Copyright (c) 2019 IEEE. Personal use of this material is permitted.  Permission from IEEE must be obtained for all other uses, in any current or future media, including reprinting/republishing this material for advertising or promotional purposes, creating new collective works, for resale or redistribution to servers or lists, or reuse of any copyrighted component of this work in other works.}
}

\maketitle

\begin{abstract}
Dense prediction models are widely used for image segmentation. One important challenge is to sufficiently train these models to yield good generalizations for hard-to-learn pixels, correct prediction of which may greatly affects the success. A typical group of such hard-to-learn pixels are boundaries between instances. Many studies have developed strategies to give specific attention to learning these boundary pixels. They include designing multi-task networks with an additional task of boundary prediction and increasing the weights of boundary pixels' predictions in the loss function. Such strategies require defining what to attend beforehand and incorporating this defined attention to the learning model. However, there may exist other groups of hard-to-learn pixels and manually defining and incorporating the appropriate attention for each group may not be feasible. In order to provide a more attainable and scalable solution, this paper proposes \textit{AttentionBoost}, which is a new multi-attention learning model based on adaptive boosting. \textit{AttentionBoost} designs a multi-stage network and introduces a new loss adjustment mechanism for a dense prediction model to adaptively learn what to attend at each stage directly on image data without necessitating any prior definition about what to attend. This mechanism modulates the attention of each stage to correct the mistakes of previous stages, by adjusting the loss weight of each pixel prediction separately with respect to how accurate the previous stages are on this pixel. This mechanism enables \textit{AttentionBoost} to learn different attentions for different pixels at the same stage, according to difficulty of learning these pixels, as well as multiple attentions for the same pixel at different stages, according to confidence of these stages on their predictions for this pixel. Using gland segmentation in histopathological images as a showcase application, our experiments demonstrate that the proposed \textit{AttentionBoost} model improves the segmentation results of its counterparts.
\end{abstract}


%\IEEEpeerreviewmaketitle

\thispagestyle{fancy}
%\lhead{}\chead{This work has been submitted to the IEEE for possible publication. Copyright may be transferred without notice, after which this version may no longer be accessible.}\rhead{}

\chead{This work has been submitted to the IEEE for possible publication. Copyright may be \\transferred without notice, after which this version may no longer be accessible.}
\renewcommand{\headrulewidth}{0.0pt}

\section{Introduction}

\IEEEPARstart{D}{ue} to their ability to learn high-level complex features on image data~\cite{zeiler14}, convolutional neural networks (CNNs) have shown a huge success on various image classification~\cite{krizhevsky12,simonyan14,szegedy15} and object detection~\cite{girshick14} tasks over the last years. For the segmentation tasks, especially dense prediction models using fully convolutional networks (FCNs) have provided significant improvements in terms of both efficiency and accuracy~\cite{long15}. Thus, FCNs have become a popular architectural choice also for medical image segmentation~\cite{litjens17}. In spite of the success of the FCNs trained on very large datasets, training may become much more difficult when small quantities of annotated data are available and when pixels of background and foreground classes are highly imbalanced, which are indeed very typical cases for medical images. In such cases, without further adjustments, the networks tend to yield poor generalizations for pixels of a minority class as well as for hard-to-learn pixels. 

The most common approach to mitigate the class-imbalance problem is to increase the relative weight of minority class predictions in the loss function. Although this approach forces the network to give more attention to learning the minority class, it may not increase the performance on hard-to-learn pixels when these pixels occur in both the majority and minority classes and when they distribute unevenly in a particular class. For instance, for the task of segmenting glands in a histopathological image, it is harder to learn the pixels close to gland boundaries, regardless of whether they belong to the foreground or the background class. Furthermore, although the number of such hard-to-learn pixels (and as a result, the total weight contribution of their predictions to the loss function) is relatively low, their correct classification greatly affects the success of the entire segmentation task since these boundary pixels separate multiple gland instances from each other.

\begin{figure}
\centering
\small{
\begin{tabular}{@{~}c@{~}c@{~}c@{~}}
\includegraphics[width =0.3\columnwidth]{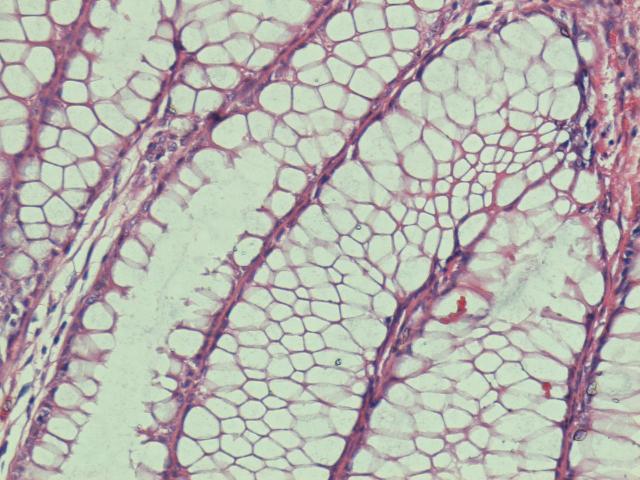} &
\includegraphics[width =0.3\columnwidth]{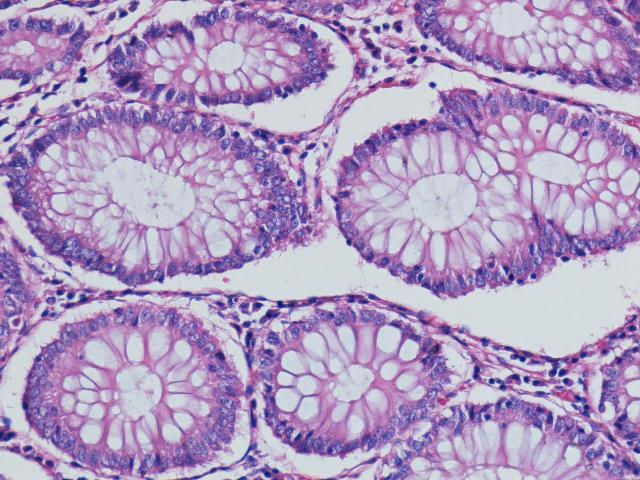} &
\includegraphics[width =0.3\columnwidth]{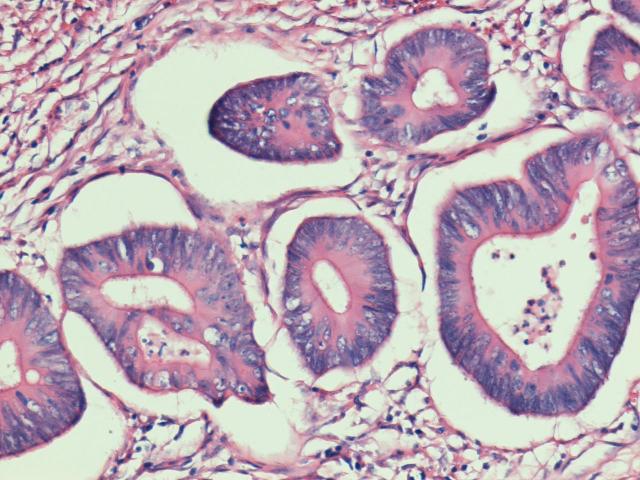} \\
(a) & (b) & (c) \\ 
\end{tabular}
}
\caption{Examples of histopathological images of colon glands. In the gland segmentation task, it is more difficult to correctly classify the boundary pixels when two glands are very close to each other. The image shown in (a) contains such kind of glands. Additionally, these images typically contain noise and artifacts due to the tissue preparation procedures. For example, due to the density difference between glands and connective tissues (inside and outside of a gland), the fixation and sectioning procedures may result in large white artifacts outside the glands. The images given in (b) and (c) contain such kind of artifacts. It is common for gland segmentation algorithms to identify some of these large white artifacts as false glands. These are the images consisting of (a)-(b) normal glands and (c) cancerous glands.}
\label{fig:motivation}
\end{figure}

To address this problem, it has been proposed to give specific attention to the classification of boundary pixels. One proposed solution is to adjust the weights of these pixels in the loss function based on their distances to the boundary of the closest gland instances~\cite{ronneberger15}. The other solution is to give this attention via designing a multi-task architecture. This has been achieved by defining boundary prediction as an additional task, learning it together with the main task of gland segmentation, and combining the predicted maps at the end, either with a simple fusion function~\cite{chen17} or with an additional fusion network~\cite{xu16}. The multi-task architecture proposed by~\cite{xu17} also includes one more additional task to predict the bounding boxes of the gland instances. Both of these solutions help better classify the boundary pixels since they give specific attention to decreasing mistakes that their network would make on these pixels. This attention is defined to alleviate one single mistake type relating to one group of hard-to-learn pixels, namely ``incorrect boundary classification'', and this mistake type needs to be manually (externally) identified before designing and learning a network. This manual identification is indeed a natural choice for the gland segmentation task since multiple gland instances may seem as touching in histopathological images due to their nature. On the other hand, there may exist other groups of hard-to-learn pixels, and thus other types of mistakes associated with these pixels, in the images (see Fig.~\ref{fig:motivation}). In order for these solutions to be scalable against multiple mistake types, either new weight adjustments or new additional tasks should be defined for each mistake type separately. Nevertheless, this should be done externally and manually, which might be challenging especially when these mistakes are not related with the nature of the images but with noise and artifacts. As shown in Fig.~\ref{fig:motivation}, histopathological images typically contain such noise and artifacts due to the tissue preparation (fixation, sectioning, and staining) procedures.

In response to these issues, this paper introduces an iterative attention learning model based on adaptive boosting. This model, which we call \textit{AttentionBoost}, proposes to learn multiple attentions directly on image data at the same time as it learns the network weights. To this end, \textit{AttentionBoost} first designs a multi-stage system that contains a fully convolutional segmentation network in each stage. Then, it proposes to modulate the attention of each segmentation network for each training image, based on the pixel-wise errors of the previous stage networks, by introducing a new loss adjustment method for a dense prediction model. This method is inspired by the Adaboost algorithm~\cite{freund97} and adjusts the loss weight of each pixel prediction separately with respect to how confident the previous stage networks are on their correct/incorrect predictions for the same pixel. By doing so, the proposed \textit{AttentionBoost} model enables to assign different attention levels to different pixels of the same image, according to the difficulty level of learning these pixels, as well as to adaptively select/learn what image parts (e.g., gland boundaries and artifacts) need more attention during the network training. This also forces the next stages to give more attention to learning the pixels incorrectly segmented by the previous stage networks. With this adaptive loss adjustment, \textit{AttentionBoost} end-to-end trains its multi-stage network and combines the outputs of all stages to obtain the final segmentation. Using gland instance segmentation as a showcase application, our experiments demonstrate that this type of attention learning improves segmentation results not only for the boundary pixels but also for other hard-to-learn pixels, mostly corresponding to false positives emerged as a result of noise and artifacts. 

\section{Related Work}

The proposed \textit{AttentionBoost} model mainly differs from the related networks in the following aspects: The literature contains single attention models that externally define what to attend before the network training starts~\cite{ronneberger15,chen17,xu16}. These attention points are manually determined as boundary pixels, assuming that these pixels are hard to learn. On the other hand, \textit{AttentionBoost} is an error-driven multi-attention model and adaptively learns what to attend directly on image data without making any prior assumption. 

\textit{AttentionBoost} is also different than the iterative methods that have been proposed to correct the mistakes of a single model and refine its results. The basic idea of these methods is to decompose a segmentation task into iterative stages where image features are learned together with high-level context features from the previous map to improve the result at the current stage~\cite{tu10,li16,shen17,gidaris17,romero17}. For that, these methods give an input image and a predicted label map from the previous stage to the next stage iteratively, starting with a null label map~\cite{li16,shen17} or a segmentation map obtained from another model~\cite{gidaris17,romero17}, and use the last predicted map after some number of iterations. As opposed to the proposed \textit{AttentionBoost} model, these methods learn the same task and use the same objective (loss) function in every stage, which does not explicitly force the network to change its attention to learning incorrectly segmented pixels but expects the network to implicitly learn how to correct its mistakes. On the other hand, although \textit{AttentionBoost} uses the same segmentation task definition in all stages, since it adaptively changes the objective function from one stage to another, it can be considered that \textit{AttentionBoost} learns a different subtask in each of these stages.

The literature also consists of studies that use different weight contributions in their loss functions. However, almost all of these studies address the class-imbalance problem. To this end, they calculate a constant weight for each class, typically inversely proportional to its pixel frequency, and use this constant weight for all predictions of the pixels belonging to the same class~\cite{eigen15,badrinarayanan17,sudre17}. Different than these studies, instead of just calculating such constant weights based on the class pixel frequencies, \textit{AttentionBoost} learns how to adjust the weights in the loss function on image data with the ability to give different weights for the pixel-wise predictions of the same class. There exists only a single study that attempts to learn the loss weights on image data for object detection~\cite{lin17}. However, this previous study neither constructs multiple networks nor trains them iteratively, but it rather focuses on training a single stage network. Each epoch of this training updates the loss weight for each object to be detected separately and the next epoch uses the same updated weight for all pixels in the bounding box of the same object. Such an approach may increase the importance of learning misdetected and most probably harder-to-learn objects in the later epochs. However, since the use of a single network requires using the same network weights for all types of object detections and since the common type of (in)correctly detected objects may still dominate the loss function, this makes harder to explicitly focus on multiple detection subtasks with different difficulty levels at the same time. On the other hand, the proposed \textit{AttentionBoost} model enables to define multiple stages, each of which can contain a network with different attention (by adaptively changing the loss function). This, in turn, allows each stage to focus on a different aspect of the segmentation task. Additionally, this previous study~\cite{lin17} uses the same loss weight for all pixels of the same object (bounding box) without considering their pixel-wise contributions. On the contrary, \textit{AttentionBoost} updates the loss weight for each pixel separately, according to the difficulty of learning this pixel.

In the literature, there also exist studies that combine the Adaboost algorithm~\cite{freund97} with a neural network architecture~\cite{schwenk00,medera09,gao16,wang16,han16}. However, these studies do not involve a dense prediction task using an FCN, but they rather focus on the task of classifying an image instance. Therefore, they use the same attention for each image by either arranging different training sets for each learner or arranging loss weights for the training instances of each learner. These non-dense prediction models, which have been designed for a classification task, are beyond the scope of this paper. This paper uses the idea to adjust the loss weights of pixel-wise predictions in a dense prediction model for a segmentation task.

\section{Methodology}

The \textit{AttentionBoost} model proposes to train a multi-stage network that adjusts (learns) the attention of each of its stages automatically and to combine the outputs of all these stages for obtaining a final segmentation. To this end, it introduces an attention learning mechanism for a dense prediction model. This mechanism relies on devising a new loss adjustment method, in which the loss contribution of each pixel prediction at each stage is adjusted depending on the confidence levels of the correct/incorrect predictions of the previous stages.

The motivation behind designing such a multi-stage network is as follows: A network is trained as to optimize its objective function, and thus, the definition of this function greatly affects the network's outputs. When there exist imbalanced data distributions and when all data points contribute to the objective function evenly, the network is biased to learning the most common patterns in the data. In this case, learning less common patterns will require adjustments in the objective function. However, making adjustments for many different patterns may not be that easy for a model that trains a single network with a single objective function. On the other hand, when the model allows training multiple (sub)networks that may use different objective (loss) functions, it is easier to make such adjustments since this gives the model an opportunity to modulate each network's attention to a different goal. 

\begin{figure}
\centering \includegraphics[width=0.85\columnwidth]{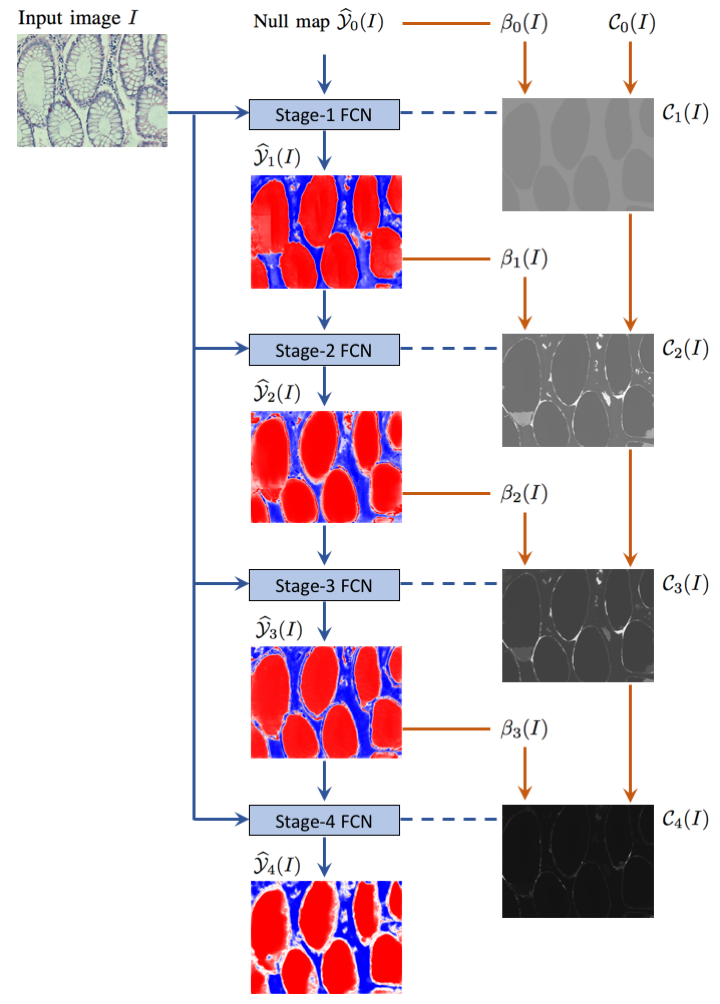}
\caption{An overview of the proposed multi-stage network architecture that consists of four segmentation networks (FCNs). The $n$-th stage network inputs an original image $I$ and a probability map $\widehat{\cal Y}_{n-1}(I)$ estimated by the previous stage and outputs a new probability map $\widehat{\cal Y}_{n}(I)$ for the next stage. While end-to-end training the multi-stage network, the loss contribution map ${\cal C}_{n}(I)$ for the $n$-th stage is modulated by $\widehat{\cal Y}_{n-1}(I)$ and ${\cal C}_{n-1}(I)$, as given in Eqns.~\ref{eqn:C} and~\ref{eqn:beta}. In order to illustrate how this multi-stage network iteratively corrects its errors for an unseen image, this figure shows the posterior maps $\widehat{\cal Y}_{n}(I)$ and loss contribution maps ${\cal C}_{n}(I)$ calculated for a test set image. Note that the loss contribution maps ${\cal C}_{n}(I)$ of this test set image are calculated just for a demonstration purpose since these maps are only calculated for the training images during the network training. In the illustration of the contribution maps, the whiter the color of a pixel is, the higher it contributes to the corresponding loss function. The posterior maps include the probability of each pixel belonging to the foreground object. In these maps, posteriors between 1 and 0.5 are shown with increasing tints of red and posteriors between 0 and 0.5 are shown with increasing tints of blue; posteriors close to 0.5 seem whitish.}
\label{fig:model}
\end{figure}

With this motivation, this paper designs a multi-stage network architecture, each stage of which trains a network with a different loss function. To do so, it iteratively takes an image and a probability map from the previous stage as the input, adjusts its loss function according to this probability map, and outputs a new probability map for the next stage. The architecture of this multi-stage network is illustrated in Fig.~\ref{fig:model} and its details are given in the following subsections.
% The source codes of its implementation are available at http://www.cs.bilkent.edu. tr/$\scriptstyle\mathtt{\sim}$gunduz/downloads/AttentionBoost.
%
%

\subsection{Attention Learning}
\label{sec:loss}

Let $I$ be an image in the training set ${\cal D}$, $p$ be a pixel in the training image $I$, and $y(p)$ be the ground truth for this pixel. Here $y(p) = 1$ if the pixel belongs to a foreground object and $y(p) = 0$ otherwise. Then, the loss function ${\cal L}_{n}$ for the $n$-th stage network is defined as
\begin{equation}
{\cal L}_{n} = \sum_{I \in {\cal D}} \sum_{p \in I} C_{n}(p) \cdot \Big( y(p) - {\hat{y}_{n}}(p) \Big) ^2
\end{equation}
where ${\hat{y}_{n}}(p)$ is the foreground probability for pixel $p$ estimated by the $n$-th stage network and $C_{n}(p)$ is the contribution of this pixel prediction to the loss function ${\cal L}_{n}$. The attention learning mechanism of the \textit{AttentionBoost} model proposes to iteratively learn these contributions $C_{n}(p)$, for each pixel $p$ and for each stage $n$, at the same time as learning the network weights by backpropagation. In particular, this mechanism decreases the loss contributions for correctly estimated pixels and increases them for incorrectly estimated ones, in the framework of adaptive boosting. 

To this end, it defines the ${\beta_{n}}(p)$ coefficient that controls how much to update the current loss contribution $C_{n}(p)$ for the next stage. That is, this coefficient is used to calculate $C_{n+1}(p)$ as follows, provided that the initial loss contributions $C_{0}(p)$ are selected with respect to the class pixel frequencies. Note that one may also select $C_{0}(p)$ the same for all pixels.
\begin{equation}
C_{n+1}(p) =  \beta_{n}(p) \cdot C_{n}(p)
\label{eqn:C}
\end{equation}

\begin{equation}
{\beta_{n}}(p) = \left\{
\begin{array}{@{~}l@{~~~~}l}
{1 - | \hat{y}_{n}(p) - 0.5 | }		& \mbox{if}~\hat{y}_{n}(p)~\mbox{is correct} \vspace{0.1in} \\ 
{1 + | \hat{y}_{n}(p) - 0.5 | }		& \mbox{if}~\hat{y}_{n}(p)~\mbox{is incorrect} \\ 
\end{array}
\right.
\label{eqn:beta}
\end{equation}

The $| \hat{y}_{n}(p) - 0.5 |$ term in Eqn.~\ref{eqn:beta} quantifies how confident the $n$-th stage network is on its estimation for pixel $p$.  Since $0 \leq | \hat{y}_{n}(p) - 0.5 | \leq 0.5$, the resulting $\beta_{n}(p)$ coefficient will converge to its minimum value of $0.5$ if the current network correctly estimates pixel $p$ and if it is very confident on this correct estimation. In this case, the loss contribution $C_{n+1}(p)$ becomes smaller, which forces the next stage network to decrease its attention to learning this pixel $p$. On the other hand, if the current network incorrectly estimates $p$ but if it is very confident on this incorrect estimation, $\beta_{n}(p)$ will converge to its maximum value of $1.5$. This time, the loss contribution $C_{n+1}(p)$ becomes larger, which forces the next stage network to increase its attention to learning pixel $p$. Thus, the $\beta_{n}(p)$ coefficients, which are calculated based on the estimations of the current stage network, are used to modulate the attention of the next stage network. 

Here it is worth to noting that after calculating the loss contributions ${C_{n+1}}(p)$ using Eqn.~\ref{eqn:C}, these contributions are normalized for the correctly estimated pixels of a training image $I$ and its incorrectly estimated pixels separately, such that $\sum {C_{n+1}}(p) = 1$ for all correctly estimated pixels $p \in I$ and $\sum {C_{n+1}}(q) = 1$ for all incorrectly estimated pixels $q \in I$. This allows the next stage networks not to completely give up their attentions to learning the correctly segmented pixels. This is important since the output maps of all stages will be aggregated at the end to obtain the final segmentation (Sec.~\ref{sec:inference}). 

\subsection{Base Model for Each Stage}
This work uses the same FCN architecture for the networks in all of its stages\footnote{\textit{AttentionBoost} does not require all networks to be the same. However, we select the same architecture for all networks for the sake of simplicity.}. The FCN at the $n$-th stage takes a normalized RGB image $I$ as an input together with the probability map $\widehat{\cal Y}_{n-1}(I) = \{ \hat{y}_{n-1}(p)\}_{p \in I}$ that is estimated for this image by the previous stage network and outputs the probability map $\widehat{\cal Y}_{n}(I) = \{\hat{y}_{n}(p)\}_{p \in I}$. In order to employ the same base model for all stages, a null map is used for $\widehat{\cal Y}_{0}(I)$ where $\hat{y}_{0}(p) = 0.5$ for all pixels. 

\begin{figure*}
\centering \includegraphics[width=0.9\textwidth]{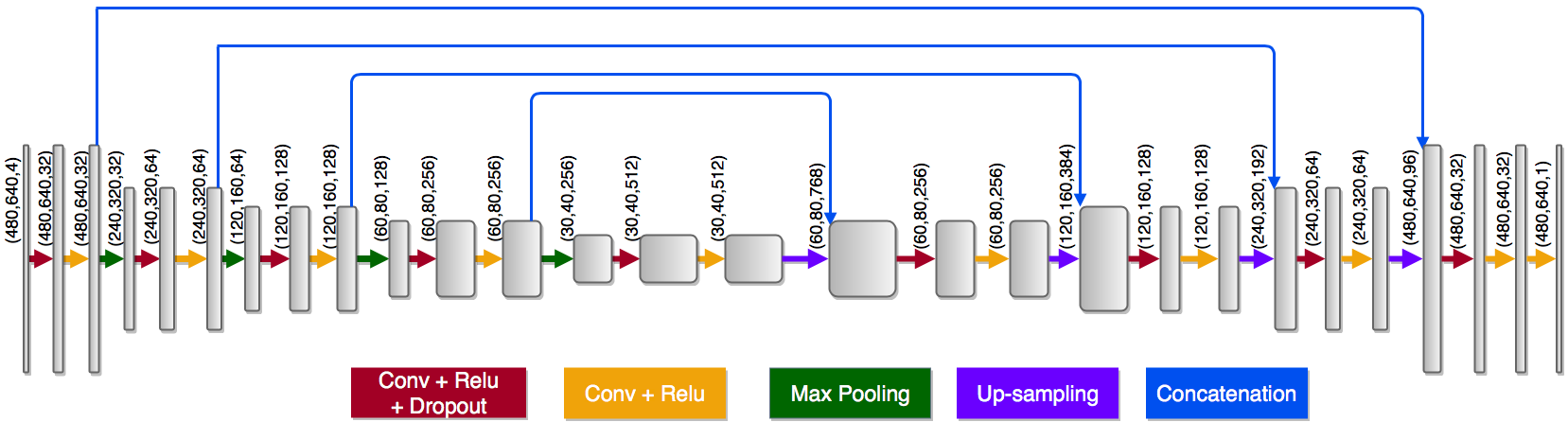}
\caption{Architecture of the FCN used as the base model. This architecture consists of an encoder and a decoder path that are connected by symmetric connections, similar to~\cite{ronneberger15}. Each box represents a feature map with its dimensions and number of channels being indicated in order on its right. Each arrow corresponds to an operation which is distinguishable by its color.}
\label{fig:basemodel}
\end{figure*}

The FCN architecture used as the base model consists of an encoder and a decoder path that are connected by symmetric connections (see Fig.~\ref{fig:basemodel}). This architecture is similar to the one proposed in~\cite{ronneberger15} where extra dropout layers~\cite{srivastava14} are added to reduce overfitting. This base model has the convolution layers with $3 \times 3$ filters and pooling/upsampling layers with $2 \times 2$ filters. It uses the sigmoid activation function at its last layer and the ReLu activation function elsewhere. Note that by using this model, our multi-stage network is fit on the memory of the GPU during end-to-end training of its four networks, and thus, its training takes faster.

\subsection{Multi-Stage Network Training}
During the network training, the normalized RGB images $I$ in the training set ${\cal D}$ are fed to the network together with their ground truth segmentation maps ${\cal Y}(I) = \{ {y}(p)\}_{p \in I}$ and the overall multi-stage network is trained in an end-to-end manner using the backpropagation algorithm. At each epoch, the forward pass calculates the loss contributions ${\cal C}_n(I) = \{ {C}_n(p)\}_{p \in I}$ for each training image $I$ from the first stage to the last one iteratively, as described in Sec.~\ref{sec:loss}. Then, the loss functions ${\cal L}_{n}$ are updated according to the calculated loss contributions and the backward pass updates the network weights by differentiating the updated loss functions. 

\subsection{Gland Segmentation}
\label{sec:inference}

After training its multi-stage network, for a given image $I$, the \textit{AttentionBoost} model aggregates the probability maps estimated by all of the stages by taking their average. Then, it first identifies the ``certain'' foreground and background regions on this average map $\widehat{\cal Y}_{avg}(I) = \{\hat{y}_{avg}(p)\}_{p \in I}$ and grows these regions onto the ``uncertain'' pixels. Here we use such an approach to alleviate the negative effects of noisy pixels that may arise in the average map due to the aggregation. This approach first classifies each pixel $p$ with a label $l(p)$ as follows, based on its average probability $\hat{y}_{avg}(p)$ and a confidence parameter $\alpha$.
\begin{equation}
{l}(p) = \left\{
\begin{array}{@{~}l@{~~~~}l}
\mbox{foreground}		& \mbox{if}~\hat{y}_{avg}(p) \ge 0.5 + \alpha \vspace{0.1in} \\ 
\mbox{background}		& \mbox{if}~\hat{y}_{avg}(p) \le 0.5 - \alpha \vspace{0.1in} \\ 
\mbox{uncertain}		& \mbox{otherwise}\\ 
\end{array}
\right.
\label{eqn:aggregate}
\end{equation}

Then, it identifies foreground and background seed regions by finding connected components of the foreground pixels and the background pixels, separately. After eliminating the seeds smaller than an area threshold $A_{thr}$ and assigning the pixels of these eliminated seeds to the uncertain class, it grows the remaining ones onto the uncertain pixels with respect to their average probabilities. Each grown foreground seed region is considered as a gland in the final segmentation map. At the end, to smooth their boundaries, a majority filter with a size of $f_{size}$ is applied on the segmented glands.

Here we use a simple approach that calculates the average over the probability maps of all stages and then uses a region growing algorithm on this average map. One may consider designing and using more sophisticated approaches to process these probability maps. This can be considered as future research work of this study.

%%%%%%%%%%%%%%%%%%%%%%%%%%%%%%%%%%%%%%%%%%%%%%%%%
\section{Experiments}
\label{sec:experiments}

\subsection{Dataset}
\par
We test our model on a dataset of 200 microscopic images of colon biopsy samples obtained from the Pathology Department Archives of Hacettepe University School of Medicine. These samples are hematoxylin-and-eosin stained tissue sections containing normal and cancerous (colon adenocarcinomatous) glands. Their images are taken using a Nikon Coolscope Digital Microscope with a $20 \times$ objective lens. The image resolution is $480 \times 640$. 

The dataset is divided into training, validation, and test sets. The training images are used by the backpropagation algorithm to learn the weights of the proposed multi-stage network and the validation images are used for early stopping of the backpropagation algorithm. Both the training and validation images are employed to select the confidence parameter $\alpha$, the area threshold $A_{thr}$, and the majority filter size $f_{size}$ used by the gland segmentation step. This parameter selection is explained in Sec.~\ref{sec:parameter-selection}. The test images are used neither for network training nor for parameter selection; they are used only for the evaluation purpose. Table~\ref{table:datasets} presents the number of images and the number of glands for each set.

\subsection{Implementation Details}

The multi-stage network containing four FCNs is implemented in Python using the Keras deep learning framework. The network is trained on the GPU (GeForce GTX 1080 Ti). It is trained from scratch using randomly initialized network weights and with an early stopping approach based on the loss calculated for the validation images. The batch size is 1 and the drop-out factor is 0.2. The learning rate and the momentum value are adaptively adjusted using the AdaDelta optimizer~\cite{zeiler12}.

\begin{table}[t]
\caption{Number of images and number of glands in the training, validation, and test sets.}
\label{table:datasets}
\centering
\setlength{\tabcolsep}{2pt}
\begin{tabular}{|l|c|c|c|c|c|c|}
\hline
& \multicolumn{3}{c|}{Number of images} &  \multicolumn{3}{c|}{Number of glands} \\ \hline
				& Training & Validation & ~~~Test~~~ & Training & Validation & ~~~Test~~~	\\ \hline

Normal 	 & 40 &10 & 50 & 570 & 174 & 621\\ \hline
Cancerous & 40 & 10 & 50 & 321 & 49 & 367\\ \hline
\rowcolor[gray]{0.75}
\textbf{\textit{Total}} 		 & 80 & 20 & 100 & 891 & 223 & 988\\ \hline
\end{tabular}
\end{table}

\subsection{Evaluation}

Segmentation results are quantitatively assessed using three criteria: 1) the object-level F-score to assess what percentage of gland objects are detected correctly, 2) the object-level Dice index to assess how accurately the pixels of the segmented gland objects overlap with those of their matching (maximally overlapping) ground truth objects, and 3) the Hausdorff distance to assess the shape similarity between the segmented gland objects and their matching ground truth objects. Note that these measures were also used in the GlaS Challenge Contest~\cite{glas}.

\subsubsection{F-score}

A segmented gland object is considered as true positive (TP) if it intersects with at least 50 percent of a ground truth object, and as false positive (FP) otherwise. A ground truth object is considered as false negative (FN) if at least its 50 percent does not intersect with any segmented gland object. The object-level F-score is defined as:
\begin{eqnarray}
\mbox{\textit{F-score}} &=& \frac{2 \cdot \mbox{\textit{precision}} \cdot \mbox{\textit{recall}}}{\mbox{\textit{precision}} + \mbox{\textit{recall}}} \\
\mbox{\textit{precision}} &=& |TP| / (|TP| + |FP|) \nonumber \\
\mbox{\textit{recall}} &=& |TP| / (|TP| + |FN|) \nonumber
\end{eqnarray}

\subsubsection{Dice index}

Let $S = \{s_i\}$ be a set of segmented gland objects in all images of a given dataset and $G = \{g_j\}$ be a set of ground truth objects in these images. To calculate the object-level Dice index on these two sets, the objects in $S$ and $G$ are first matched: Each $s_i \in S$ is matched with a ground truth object $\gamma(s_i) \in G$ that maximally overlaps $s_i$. Similarly, each $g_j \in G$ is matched with a segmented gland object $\sigma(g_j) \in S$ that maximally overlaps $g_j$. Then, by accumulating the Dice indices calculated for all matching object pairs, the object-level Dice index is defined as follows:
\begin{equation}
\label{eqn:dice}
\mbox{\textit{Dice}}(S, G)=\frac{1}{2}\begin{pmatrix}
~\sum \limits_{s_i \in S} \omega(s_i) \cdot DI(s_i, \gamma(s_i))~ \\
+ \vspace{0.2 cm}\\ 
\sum \limits_{g_j \in G} \omega(g_j) \cdot DI(g_j, \sigma(g_j))
\end{pmatrix}
\end{equation}
where $\omega(s_i) = |s_i| ~/ \sum \limits_{s_m \in S} s_m$ and $\omega(g_j) = |g_j| ~/ \sum \limits_{g_m \in G} g_m$. Here $DI(x, y) = 2 \cdot |x \cap y| / (|x| + |y|)$ is the Dice index of a pair of objects $x$ and $y$, one from the segmented gland objects and the other from the ground truth objects.  Note that if there is no matching ground truth object of a segmented gland object (or vice versa), the contribution of this object to the Dice index is zero.

\subsubsection{Hausdorff distance}

Likewise, the objects in $S$ and $G$ are matched to calculate the object-level Hausdorff distance. Each $s_i \in S$ is matched with $\gamma(s_i) \in G$ that maximally overlaps $s_i$. If there is no overlap, $\gamma(s_i)$ is the ground truth object that has the minimum Hausdorff distance from $s_i$. Similarly, each $g_j \in G$ is matched with $\sigma(g_j) \in S$ that maximally overlaps $g_j$. If there is no overlap, $\sigma(g_j)$ is the segmented gland object that has the minimum Hausdorff distance from $g_j$. Then, by accumulating the Hausdorff distances calculated for all matching object pairs, the object-level Hausdorff distance is defined as follows:
\begin{equation}
\label{eqn:hausdorff}
\mbox{\textit{Hausdorff}}(S, G)=\frac{1}{2}\begin{pmatrix}
~\sum \limits_{s_i \in S} \omega(s_i) \cdot HD(s_i, \gamma(s_i))~ \\
+ \vspace{0.2 cm}\\ 
\sum \limits_{g_j \in G} \omega(g_j) \cdot HD(g_j, \sigma(g_j))
\end{pmatrix}
\end{equation}
$HD(x, y) = \max\{ \sup\limits_{{p_x} \in x } \inf\limits_{{p_y} \in y} ||p_x - p_y||,  \sup\limits_{{p_y} \in y } \inf\limits_{{p_x} \in x} ||p_x - p_y|| \}$ is the Hausdorff distance between a pair of objects $x$ and $y$, one from the segmented gland objects and the other from the ground truth objects. Note that $\sup\limits_{{p_x} \in x } \inf\limits_{{p_y} \in y} ||p_x - p_y||$ gives the maximum of the minimum distances calculated from every pixel $p_x$ of the object $x$ to any pixel $p_y$ of the object $y$.

\subsection{Parameter Selection}
\label{sec:parameter-selection}

\textit{AttentionBoost} uses three external parameters in its gland segmentation step.  These are the confidence parameter $\alpha$ to identify certain pixels for region growing, the area threshold $A_{thr}$ to eliminate small regions, and the majority filter size $f_{size}$ to control how much to smooth gland boundaries. The grid search is used to select their values. For that, all combinations of $\alpha = \{0.05, 0.10, 0.15, 0.20, 0.25\}$, $A_{thr} = \{250, 500, 750, 1000\}$, $f_{size} = \{5, 9, 15, 19\}$ are considered and the one that yields the highest Dice index for the training and validation images is selected. The test set images are not used in this selection at all. The selected values are $\alpha = 0.15$, $A_{thr} = 250$ and $f_{size} = 15$. Sec.~\ref{sec:parameter-analysis} will discuss the effects of this parameter selection to the model's performance in detail. Note that the same procedure is used to select the external parameters of the comparison methods. 

\subsection{Comparisons}

We compare our model with three approaches implemented based on the previously reported dense prediction models~\cite{ronneberger15,chen17,li16}. The first two, the \textit{BoundaryAttentionWithLossAdjustment} and \textit{BoundaryAttentionWithMultiTask} methods, are single-stage models that give specific attention to predicting gland boundaries. However, as opposed to our proposed model, which automatically learns multiple attentions directly on image data, these comparison methods require a prior definition of what to attend and include this definition in their system design. We use these comparison methods to explore the benefits of our proposed multi-attention learning. The last comparison method, \textit{MultiStageWithoutAdaptiveBoosting}, is a multi-stage model, each stage of which also takes an input image and a segmentation (probability) map from the previous stage and produces another segmentation map for the next stage. However, different than our model, it always uses the same objective (loss) function at all of its stages. It neither explicitly forces its network to modulate its attention to learning incorrectly predicted pixels nor employs adaptive boosting for this purpose. We use this last comparison method to understand the effectiveness of using adaptive boosting in a dense prediction model. The details of these three comparison methods are given below. Note that for fair comparisons, all these methods use the same FCN architecture, which is given in Fig.~\ref{fig:basemodel}, in their base models. 

\begin{table*}[t]
\caption{Quantitative results of the proposed \textit{AttentionBoost} model and the comparison methods obtained on the test set images.}
\label{table:results}
\centering
\setlength{\tabcolsep}{2pt}
\begin{tabular}{|l|c|c|c|c|c|c|c|c|c|}
\hline
& \multicolumn{3}{c|}{Normal glands} &  \multicolumn{3}{c|}{Cancerous glands} &  \multicolumn{3}{c|}{All glands} \\ \hline
				& ~F-score~ & ~~~Dice~~~ & Hausdorff & ~F-score~ & ~~~Dice~~~ & Hausdorff & ~F-score~ & ~~~Dice~~~ & Hausdorff \\ \hline

\textit{AttentionBoost} 	 				& 95.39	& 94.58	& 25.89	& 91.76	& 92.50	& 42.74	& 94.03	& 93.56	& 34.12 \\ \hline			
\textit{BoundaryAttentionWithLossAdjustment} 	& 89.39	& 86.36	& 71.16	& 87.57	& 90.66	& 55.09	& 88.69	& 88.46	& 63.29 \\ \hline
\textit{BoundaryAttentionWithMultiTask} 	 	& 95.59	& 92.48	& 33.51	& 84.14	& 89.84	& 46.05	& 91.13	& 91.20	& 39.61 \\ \hline
\textit{MultiStageWithoutAdaptiveBoosting} 	& 88.50	& 84.04	& 86.08	& 90.60	& 91.66	& 50.37	& 89.31	& 87.77	& 68.62 \\ \hline
\end{tabular}
\end{table*}

\subsubsection{\textit{BoundaryAttentionWithLossAdjustment}} 

It gives specific attention to learning boundary pixels by increasing the importance of their correct prediction. For that, it adjusts the loss contributions of all pixels based on their distances to the boundary of the closest gland instances, as explained in~\cite{ronneberger15}. Note that this method relies on the U-Net model that uses loss adjustments in its training. The pixels predicted as gland by this trained network typically form undersegmented components for multiple gland instances that are close to each other; some of these instances are connected to each other by narrow bridges. Thus, to improve the results of this comparison method, the gland pixels are postprocessed as follows: They are first eroded by a disk structuring element, eroded components smaller than a threshold are eliminated, and the remaining components are dilated by using the same structuring element. Here the size of the structuring element and the threshold are selected using the grid search on the training and validation images (see Sec.~\ref{sec:parameter-selection}).
\subsubsection{\textit{BoundaryAttentionWithMultiTask}}

This method gives specific attention to learning boundary pixels by designing a multi-task architecture, similar to the DCAN model proposed in~\cite{chen17}. This architecture defines an additional task for boundary prediction and concurrently learns it together with the main task of gland segmentation. After training its network, the \textit{BoundaryAttentionWithMultiTask} method locates glands in an image by subtracting the predicted boundary pixels from the predicted gland pixels and applying postprocessing. The postprocessing includes finding large connected components on the subtracted map and dilating them with a disk structuring element. Likewise, the area threshold and structuring element size are selected by the grid search.
\subsubsection{\textit{MultiStageWithoutAdaptiveBoosting}} 

It uses the same multi-stage network of the proposed \textit{AttentionBoost} model and iteratively trains this network as proposed in~\cite{li16}. However, it uses the same loss function at all of its stages and does not use adaptive boosting at all. After its training, the segmentation map produced by its last stage is taken and postprocessed to locate glands in a given image. Its postprocessing procedure is the same with that of the \textit{BoundaryAttentionWithLossAdjustment} method. The parameters used in this procedure are also selected by the grid search.
\begin{figure*}[!t]
\centering
\footnotesize{
\begin{tabular}{@{~~}c@{~~}c@{~~}c@{~~}c@{~~}c@{~~}c@{~~}}
&  &  & \bf{BoundaryAttention} & \bf{BoundaryAttention} & \bf{Multi-Stage}\\
\bf{Images} & \bf{Ground truths} & \bf{AttentionBoost} & \bf{(with loss adjustment)} & \bf{(with multi-task)} & \bf{(no adaptive boosting)} \\
\includegraphics[width =0.3\columnwidth]{gts21.jpg} & 
\includegraphics[width =0.3\columnwidth]{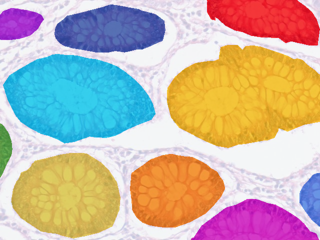} & 
\includegraphics[width =0.3\columnwidth]{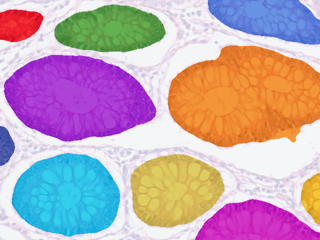} & 
\includegraphics[width =0.3\columnwidth]{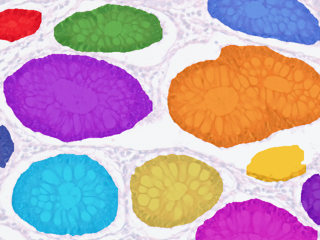} & 
\includegraphics[width =0.3\columnwidth]{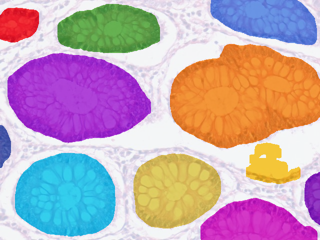} & 
\includegraphics[width =0.3\columnwidth]{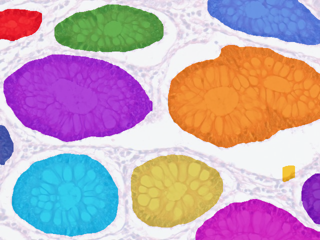}  \\
\includegraphics[width =0.3\columnwidth]{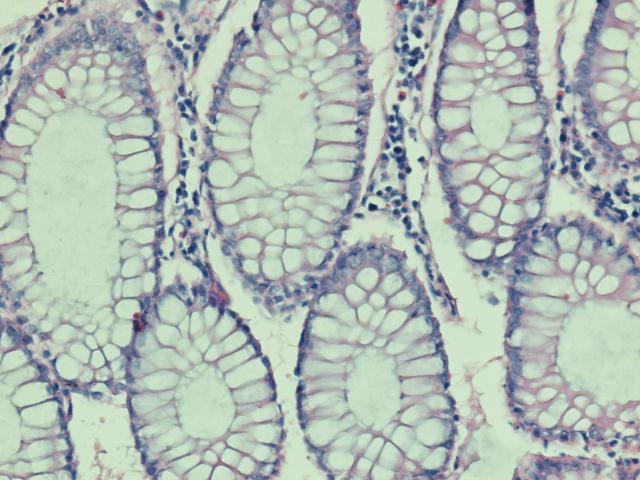} & 
\includegraphics[width =0.3\columnwidth]{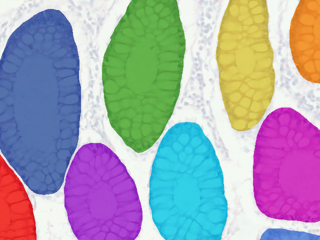} & 
\includegraphics[width =0.3\columnwidth]{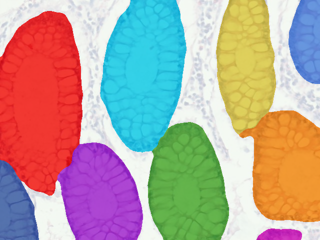} & 
\includegraphics[width =0.3\columnwidth]{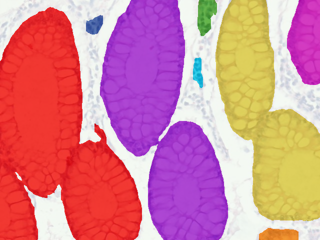} & 
\includegraphics[width =0.3\columnwidth]{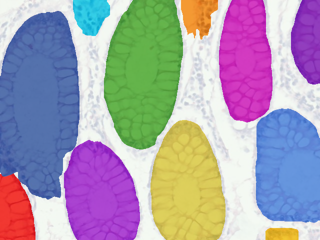} & 
\includegraphics[width =0.3\columnwidth]{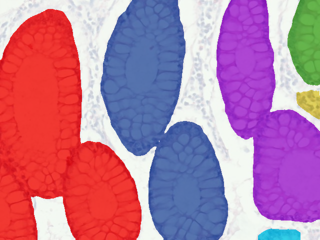}  \\
\includegraphics[width =0.3\columnwidth]{gts34.jpg} & 
\includegraphics[width =0.3\columnwidth]{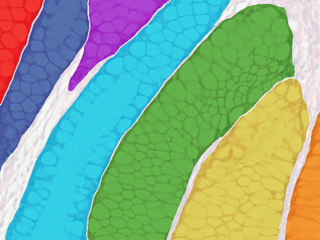} & 
\includegraphics[width =0.3\columnwidth]{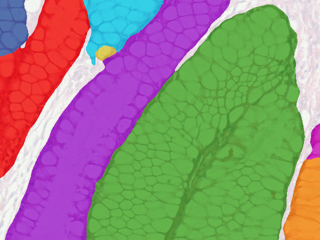} & 
\includegraphics[width =0.3\columnwidth]{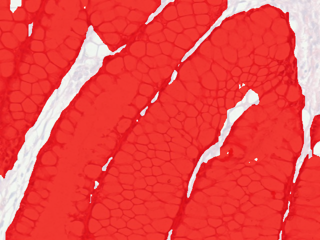} & 
\includegraphics[width =0.3\columnwidth]{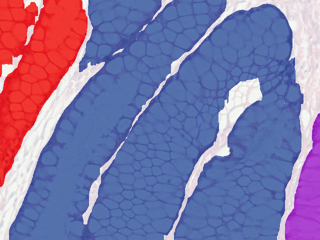} & 
\includegraphics[width =0.3\columnwidth]{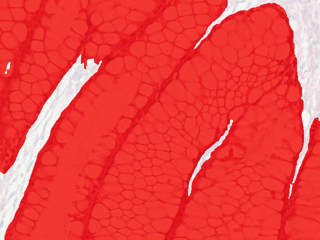}  \\
\includegraphics[width =0.3\columnwidth]{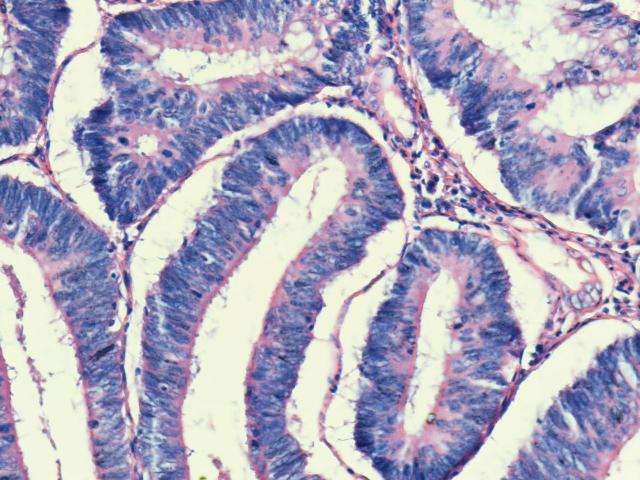} & 
\includegraphics[width =0.3\columnwidth]{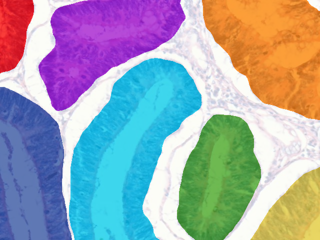} & 
\includegraphics[width =0.3\columnwidth]{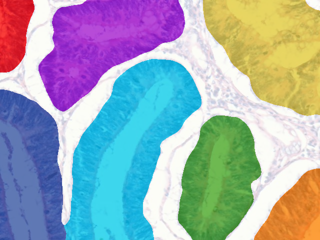} & 
\includegraphics[width =0.3\columnwidth]{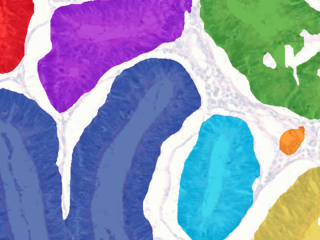} & 
\includegraphics[width =0.3\columnwidth]{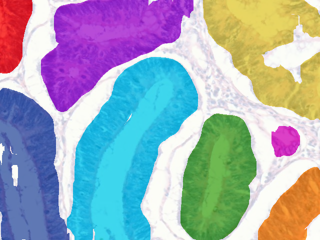} & 
\includegraphics[width =0.3\columnwidth]{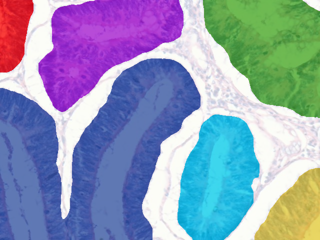}  \\
\includegraphics[width =0.3\columnwidth]{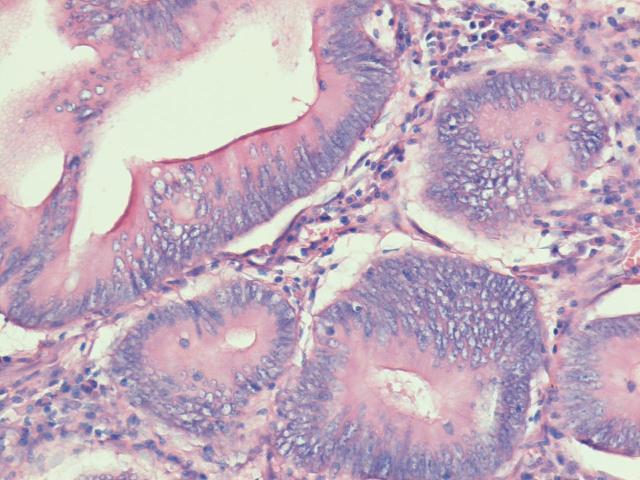} & 
\includegraphics[width =0.3\columnwidth]{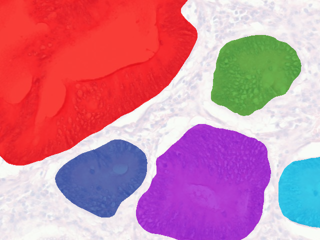} & 
\includegraphics[width =0.3\columnwidth]{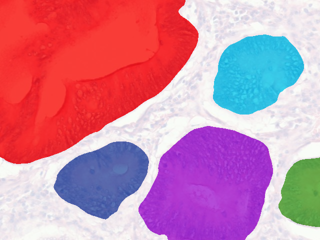} & 
\includegraphics[width =0.3\columnwidth]{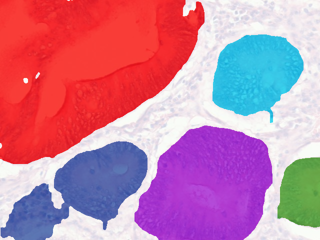} & 
\includegraphics[width =0.3\columnwidth]{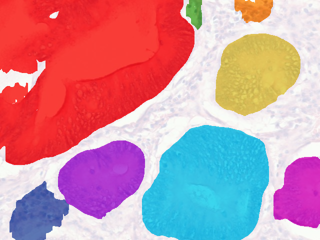} & 
\includegraphics[width =0.3\columnwidth]{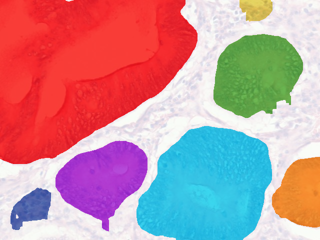}  \\
\includegraphics[width =0.3\columnwidth]{gts86.jpg} & 
\includegraphics[width =0.3\columnwidth]{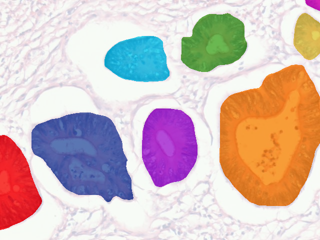} & 
\includegraphics[width =0.3\columnwidth]{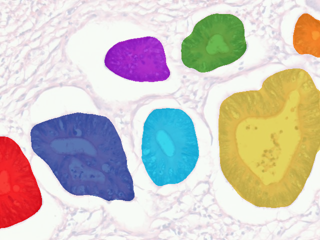} & 
\includegraphics[width =0.3\columnwidth]{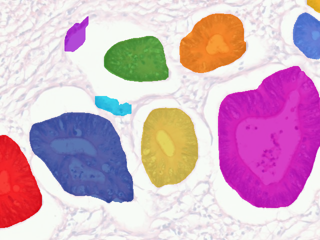} & 
\includegraphics[width =0.3\columnwidth]{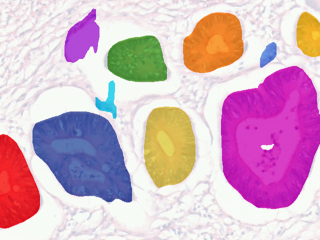} & 
\includegraphics[width =0.3\columnwidth]{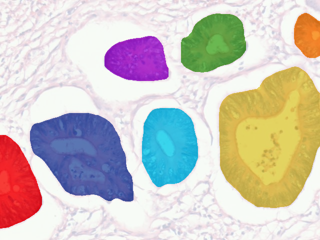}  \\
(a) & (b) & (c) & (d) & (e) & (f) \\
\end{tabular}}
\caption{(a) Example images containing normal (first three rows) and cancerous (last three rows) glands. (b) Ground truths. (c) Results of the proposed \textit{AttentionBoost} model. (d) Results of the \textit{BoundaryAttentionWithLossAdjustment} method, which gives specific attention to learning boundaries by changing the loss contributions of the boundary pixel predictions~\cite{ronneberger15}. (e) Results of the \textit{BoundaryAttentionWithMultiTask} method, which gives specific attention to learning boundaries by defining an additional task~\cite{chen17}. (f) Results of the \textit{MultiStageWithoutAdaptiveBoosting} method, which uses a multi-stage network without adaptive boosting (without learning and adaptively changing the loss contributions)~\cite{li16}. Note that these are the test set images; they are not used in any part of network training or parameter selection.}
\label{fig:visual-results}
\end{figure*}

\section{Results}

Table~\ref{table:results} reports the quantitative results of our proposed \textit{AttentionBoost} model as well as those of the comparison methods. It presents the results obtained on all of the test set images as well as those obtained on the test set images containing normal and cancerous glands, separately. These results show that \textit{AttentionBoost} is more successful at detecting and segmenting glands (higher F-score and Dice index values) as well as it yields more accurate gland shapes (lower Hausdorff distances). This is attributed to the ability of our model to automatically learn what to attend in images as well as to focus on different types of mistakes. To explore this further, we examine the following types of mistakes the methods make in their segmentations, visually (Fig.~\ref{fig:visual-results}) and quantitatively (Table~\ref{table:results-more}).
\begin{itemize}
\item \textit{Undersegmented ground truth objects:} A ground truth object $g \in G$ is considered as undersegmented if a segmented gland object $s \in S$ intersects with at least 50 percent of $g$ but also intersects with at least 50 percent of another ground truth object $g' \in G$. This mistake type commonly occurs when a method cannot correctly predict the labels of pixels close to the gland boundaries. As also mentioned in the introduction, this is the mistake type that most of the previous methods have attempted to solve by either adjusting the weights of the boundary pixels in the loss function~\cite{ronneberger15} or defining boundary prediction as an additional task in a multi-task architecture~\cite{chen17,xu16}.
\item \textit{False positives:} A segmented gland object $s \in S$ is considered as false positive if it does not intersect with at least 50 percent of any ground truth object $g \in G$. In our experiments, we observe this mistake type due to two main reasons. The first one is to segment non-gland regions as gland objects. These non-gland regions are typically located around white artifacts, which are usually formed in tissues as a result of the tissue preparation (fixation and sectioning) procedures. Such an example can be seen in the first row of Fig.~\ref{fig:visual-results}(d). The second reason is to oversegment small objects in a gland, usually close to its boundary. Two such examples (two small oversegmented objects) can be seen in the third row of Fig.~\ref{fig:visual-results}(c). To distinguish these two sorts of false positives, we call $s \in S$ a false segmented object if it does not intersect with at least 50 percent of any $g \in G$ and if any $g' \in G$ does not intersect with at least 50 percent of $s$. On the other hand, we call it a small oversegmented object, again if it does not intersect with at least 50 percent of any $g \in G$ but if a ground truth object $g' \in G$ intersects with at least 50 percent of $s$.
\item \textit{False negatives:}  A ground truth object $g \in G$ is considered as false negative (missing object) if at least its 50 percent does not intersect with any segmented gland object $s \in S$. 
\end{itemize}

\begin{table*}[t]
\caption{Number of the types of mistakes that the proposed \textit{AttentionBoost} model and the comparison methods make on the test set images.}
\label{table:results-more}
\centering
\begin{tabular}{| l | c | c | c | c |}
\hline
					& Undersegmented 		& False 				&  Small 					& Missing  \\
					& ground truth objects 	& ~~segmented objects~	&  oversegmented objects 	& ground truth objects \\ \hline

\textit{AttentionBoost} 	 				& 60		& 15		& 27 		& 42 \\ \hline			
\textit{BoundaryAttentionWithLossAdjustment} 	& 222	& 46		& 15		& 20 \\ \hline
\textit{BoundaryAttentionWithMultiTask} 	 	& 80		& 55 		& 50		& 30 \\ \hline
\textit{MultiStageWithoutAdaptiveBoosting} 	& 215  	& 16		& 16		& 31 \\ \hline
\end{tabular}
\end{table*}

\begin{figure*}[!t]
\centering
\footnotesize{
\begin{tabular}{@{~~}c@{~~}c@{~~}c@{~~}c@{~~}c@{~~}c@{~~}}
\includegraphics[width =0.3\columnwidth]{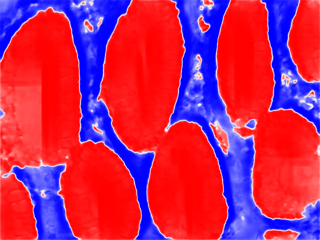} & 
\includegraphics[width =0.3\columnwidth]{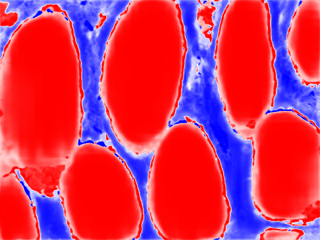} & 
\includegraphics[width =0.3\columnwidth]{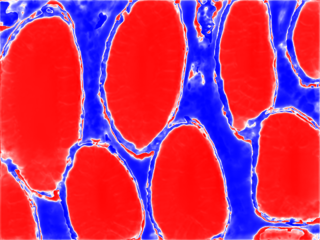} & 
\includegraphics[width =0.3\columnwidth]{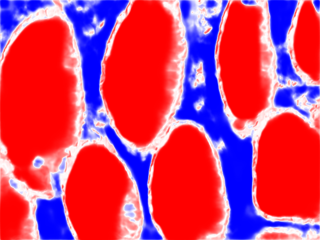} & 
\includegraphics[width =0.3\columnwidth]{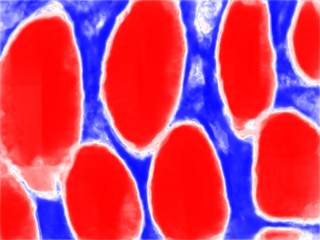} & 
\includegraphics[width =0.3\columnwidth]{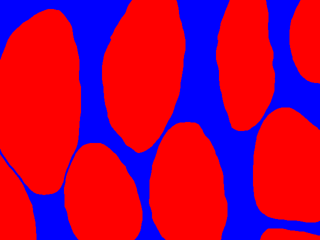} \\
\includegraphics[width =0.3\columnwidth]{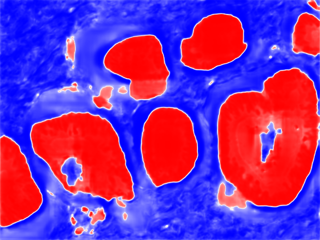} & 
\includegraphics[width =0.3\columnwidth]{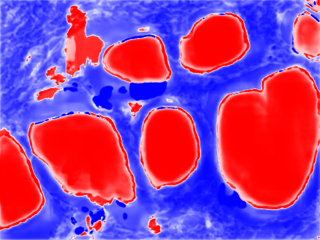} & 
\includegraphics[width =0.3\columnwidth]{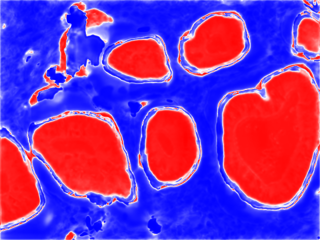} & 
\includegraphics[width =0.3\columnwidth]{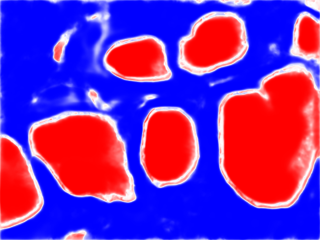} & 
\includegraphics[width =0.3\columnwidth]{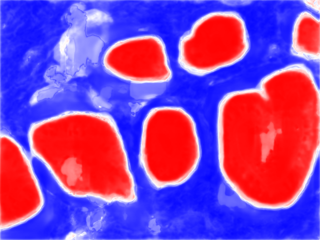} & 
\includegraphics[width =0.3\columnwidth]{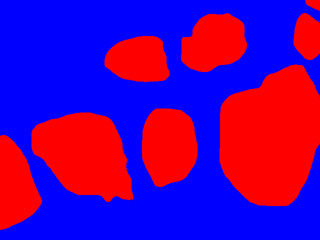} \\
(a) & (b) & (c) & (d) & (e) & (f) \\
\end{tabular}}
\caption{(a) Posterior map $\widehat{\cal Y}_{1}(I)$ generated by the first stage. (b) Posterior map $\widehat{\cal Y}_{2}(I)$ generated by the second stage. (c) Posterior map $\widehat{\cal Y}_{3}(I)$ generated by the third stage. (d) Posterior map $\widehat{\cal Y}_{4}(I)$ generated by the fourth stage. (e) Average posterior map $\widehat{\cal Y}_{avg}(I)$ obtained by aggregating the posterior maps of all stages. (f) Posterior map ${\cal Y}(I)$ produced by the ground truth segmentation. These maps include the pixel posteriors where 1 indicates that a pixel belongs to the gland class and 0 indicates that it belongs to the background. Posteriors between 1 and 0.5 are shown with increasing tints of red and posteriors between 0 and 0.5 are shown with increasing tints of blue. Note that in these images posteriors close to 0.5 seem whitish.}
\label{fig:redblue}
\end{figure*}

The number of the types of mistakes that the methods make on the test set images are reported in Table~\ref{table:results-more} and the visual results on exemplary test set images are provided in Fig.~\ref{fig:visual-results}. These results demonstrate that the proposed \textit{AttentionBoost} model leads to the best results both for undersegmentations, which emerge as a result of incorrectly classifying boundary pixels, and for false segmented objects, which are incorrectly located because of not differentiating true gland pixels from those that belong to non-gland regions mostly containing noise and artifacts. These are the two most common mistake types for this gland segmentation problem and our proposed model improves segmentation results for both at the same time, in contrast to its counterparts, which are good at either one mistake type or the other. This improvement is attributed to the following: \textit{AttentionBoost} is a multi-stage and an error-driven multi-attention learning model, each stage of which is able to give a different level of attention to learning different parts (pixels) of an image. This enables each stage to produce a segmentation (posterior) map complementary to those of the other stages. The maps of different stages are complementary on the incorrect predictions, especially for hard-to-learn pixels, since it is usually quite difficult for a single network to produce the correct predictions for all such pixels. By having such complementary maps, errors in one map may be compensated by another. Thus, when these maps are aggregated, it is expected to obtain more robust predictions. This can also be seen in Fig.~\ref{fig:redblue} that provides the posterior maps produced for two exemplary test set images. Note that \textit{AttentionBoost} misses slightly more ground truth objects. However, in our experiments, we observe that most of them correspond to small ground truth objects close to image edges. The one at the upper-right corner of the image shown in the last row of Fig.~\ref{fig:visual-results}(b) is an example of such small objects.

\begin{figure*}[!t]
\centering
\small{
\begin{tabular}{ccc}
\includegraphics[width=0.58\columnwidth]{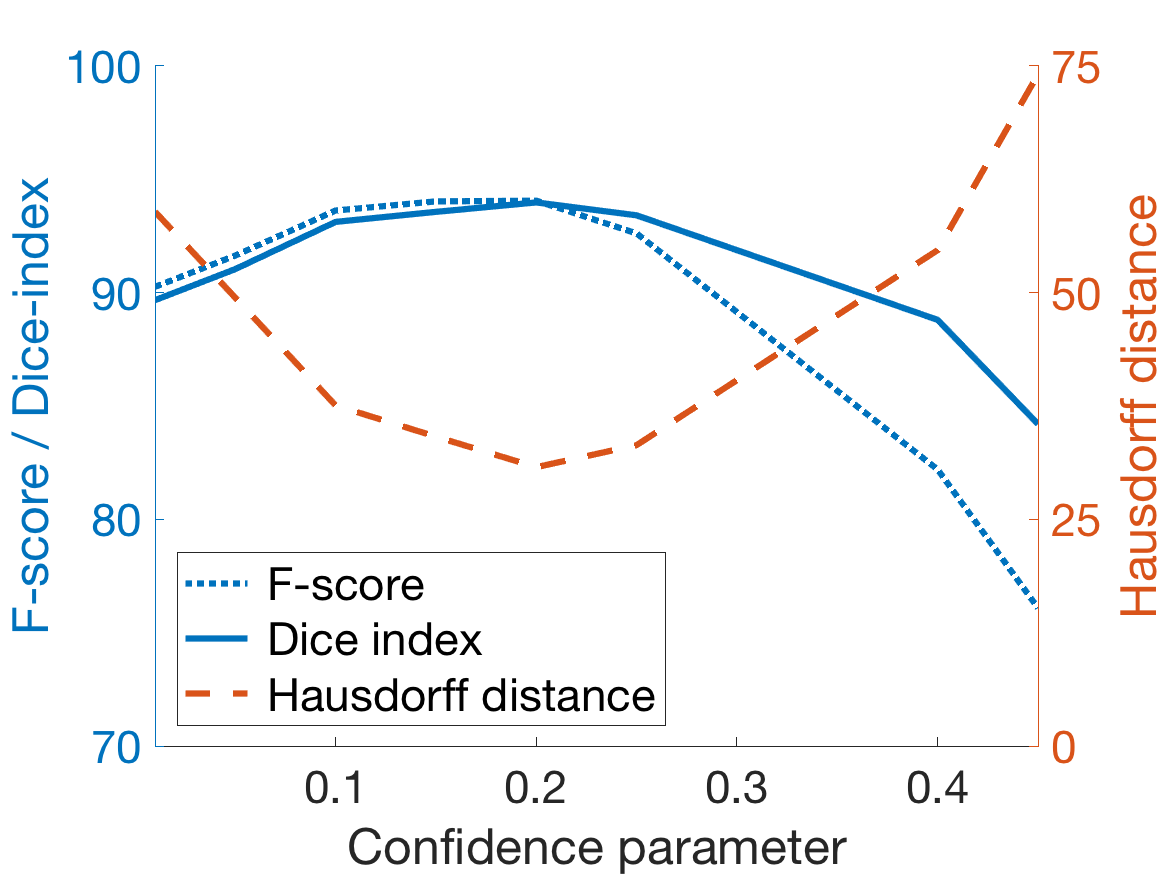} &
\includegraphics[width=0.58\columnwidth]{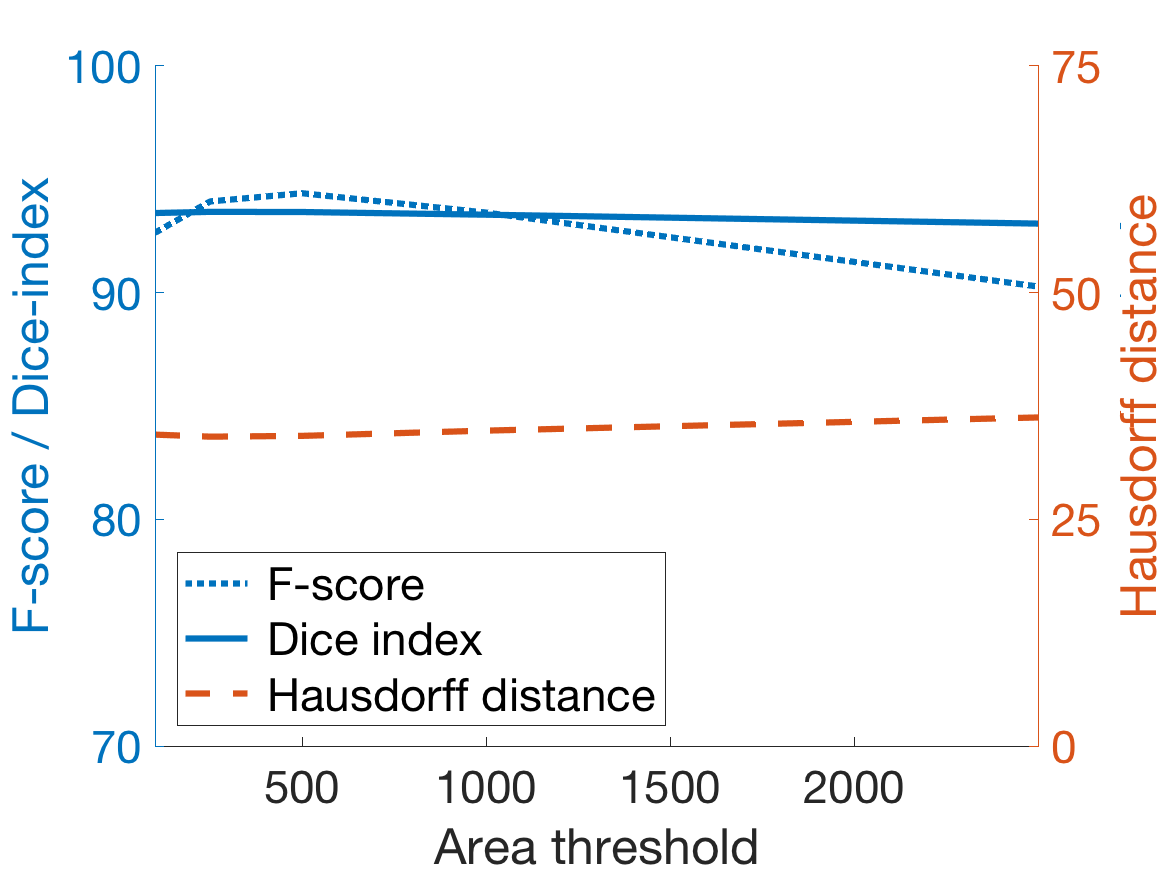} &
\includegraphics[width=0.58\columnwidth]{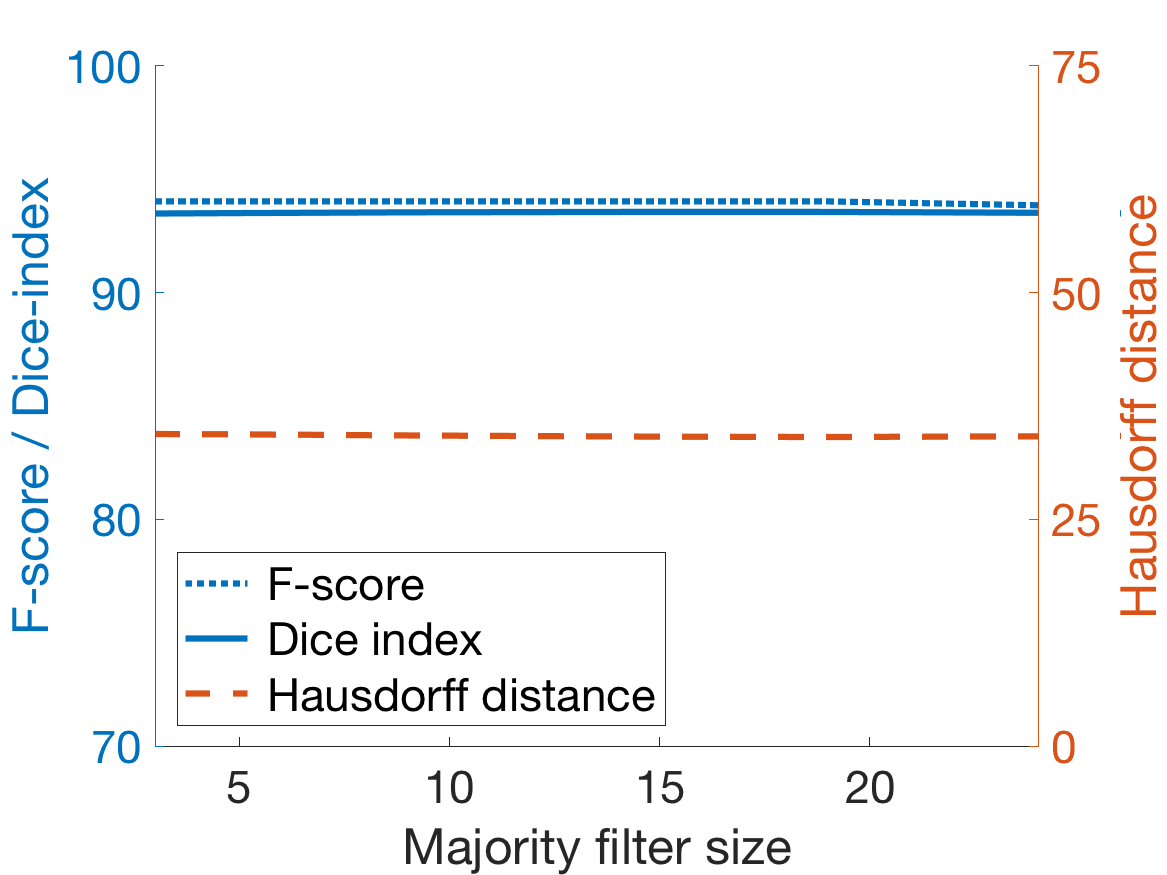}  \\
(a) & (b)  & (c) \\
\end{tabular}
}
\caption{Test set F-scores, Dice indices, and Hausdorff distances as a function of the model parameters: (a) confidence parameter $\alpha$, (b) area threshold $A_{thr}$, and (c) majority filter size $f_{size}$.}
 \label{fig:parameter-analysis}
\end{figure*}

When these results are compared with those of the other methods, we have the following observations: First, \textit{MultiStageWithoutAdaptiveBoosting}, which is also a multi-stage model but uses the same loss function in all of its stages, is successful to eliminate false positives. However, it cannot sufficiently improve boundary pixel prediction throughout its stages, which leads to a significantly higher number of undersegmentations. This suggests the benefits of automatically adjusting the loss functions of consecutive stages via adaptive boosting. Second, \textit{BoundaryAttentionWithMultiTask}, which designs a multi-task architecture that includes an additional task to give specific attention to boundary pixel prediction, gives relatively better results for undersegmentations. On the other hand, this method is effective for this specific mistake type at the expense of locating more false positives, as also seen in Fig~\ref{fig:visual-results}(e). This indicates the effectiveness of learning multiple attentions directly on image data instead of externally defining specific attention type beforehand. The proposed \textit{AttentionBoost} model adaptively learns multiple attentions by designing a multi-stage network and modulating the attention of each stage by adaptive boosting. Last, \textit{BoundaryAttentionWithLossAdjustment} is less successful for reducing both undersegmented ground truth objects and false segmented glands. Most probably, it tends to locate glands more than necessary, which also results in missing only a small number of ground truth objects.

\subsection{Parameter Analysis}
\label{sec:parameter-analysis}

\textit{AttentionBoost} has three external parameters used in its gland segmentation step: confidence parameter $\alpha$, area threshold $A_{thr}$, and filter size $f_{size}$. We analyze the effects of these parameters on the model's performance. To this end, for each parameter, we fix the selected values of the other two parameters and measure the test set F-score, Dice index, and Hausdorff distance as a function of the parameter of interest. These analyses are depicted in Fig.~\ref{fig:parameter-analysis}. 

The gland segmentation step inputs the average probability map $\widehat{\cal Y}_{avg}(I) = \{\hat{y}_{avg}(p)\}_{p \in I}$ for an image $I$ and locates gland objects on this map. For that, it first identifies certain foreground and background pixels, from which the gland objects and background are grown. The confidence parameter $\alpha$ determines which pixels are to be considered as certain, as given in Eqn.~\ref{eqn:aggregate}. When this parameter is selected too large, only the pixels $p$ for which $\hat{y}_{avg}(p)$ is very close to 1 are selected for the foreground and those for which $\hat{y}_{avg}(p)$ is very close to 0 are selected for the background. Such average posteriors can only be obtained when the networks at all stages give the same output with high confidence. However, this is not an expected output of our multi-stage network, especially for hard-to-learn pixels, since it is designed with the purpose of correcting the mistakes of one stage in another. Thus, larger $\alpha$ values result in selecting a smaller number of certain foreground pixels, which decreases the number of gland objects to be grown. This, in turn, greatly lowers the model's performance (lower F-scores, lower Dice indices, and higher Hausdorff distances). On the other hand, when this parameter is selected too small, almost all pixels are considered as certain. This also lowers the performance, by leading to more undersegmented gland objects, since pixels whose $\hat{y}_{avg}(p)$ is around 0.5 are typically found on gland boundaries and these pixels are considered as certain when smaller $\alpha$ values are used. This analysis is depicted in Fig.~\ref{fig:parameter-analysis}(a).

The area threshold $A_{thr}$ is used to eliminate small certain seed regions, from which the gland objects and the background are grown. Too small $A_{thr}$ values cannot eliminate noisy gland objects, which leads to false positives. On the other hand, too large $A_{thr}$ values also eliminate small true glands, which this time leads to false negatives. Both of them lower the F-score. Here it is worth to noting that this parameter only slightly affects the Dice index and Hausdorff distance. The reason is that: Both of these measures are weighted averages of the Dice indices and Hausdorff distances calculated on individual gland objects, where the weights are determined by the areas of these objects (see Eqns.~\ref{eqn:dice} and~\ref{eqn:hausdorff}). Since this elimination is typically applicable to small-sized glands, it does not change these measures too much. This analysis is depicted in Fig.~\ref{fig:parameter-analysis}(b).

The last parameter is the filter size $f_{size}$ of the majority filter, which is applied on the grown gland objects to smooth their boundaries. Although it improves the appearance of the glands boundaries, this parameter does not change the number of the detected glands or does not change their areas too much. Thus, it only very slightly affects the performance measures, as shown in Fig.~\ref{fig:parameter-analysis}(c).

\section{Conclusion}

This paper presents an error-driven multi-attention learning model for image segmentation. This model, which we call \textit{AttentionBoost}, relies on designing a multi-stage network and adaptively learning what image parts (pixels) each stage needs to attend and the level of this attention directly on image data. To this end, it introduces a new loss adjustment mechanism that uses adaptive boosting for a dense prediction model for the first time. This mechanism modulates the attention of each stage to correct the mistakes of its previous stages, by adjusting the loss weight of each pixel separately according to how confident the previous stages are on their predictions for this pixel. We tested our model for the problem of gland instance segmentation in histopathological images. Our experiments revealed that the proposed \textit{AttentionBoost} model, which enables to learn different attentions for different pixels at the same stage as well as to learn multiple attentions for the same pixel at different stages, leads to more accurate segmentation results compared to the existing approaches. 

For an unseen image, \textit{AttentionBoost} obtains the probability map by averaging those estimated by all stages of the multi-stage network. Then, it applies a simple seed-controlled region growing algorithm on the average map. One future research direction is to investigate more sophisticated ways of combining the probability maps of different stages. For example, one can train another neural network that inputs these probability maps and outputs the final segmentation. This work used gland instance segmentation as a showcase application. Applying this model for other instance segmentation problems is considered as another future research direction of this study.

%%%%%%%%%%%%%%%%%%%%%%%%%%%%%%%%%%%%%%%%%%%%%%%%%%%%%%%%%%%
%\section*{References}
%\bibliographystyle{IEEEtran}
%\bibliography{ref} 

%%%%%%%%%%%%%%%%%%%%%%%%%%%%%%%%%%%%%%%%%%%%%%%%%%%%%%%%%%%
\end{document}